\documentclass{article}
\pdfoutput=1

\usepackage[preprint,nonatbib]{neurips_2023}

\usepackage[utf8]{inputenc} %
\usepackage[T1]{fontenc}    %
\usepackage{hyperref}       %
\usepackage{url}            %
\usepackage{booktabs}       %
\usepackage{amsfonts}       %
\usepackage{nicefrac}       %
\usepackage{microtype}      %
\usepackage{xcolor}         %
\usepackage{xspace}
\usepackage{textcomp}
\usepackage{graphicx}
\usepackage{wrapfig}

\usepackage{array}
\usepackage{soul}
\newcolumntype{P}[1]{>{\centering\arraybackslash}p{#1}}
\usepackage{float}
\usepackage{pifont}%
\usepackage{afterpage}
\usepackage{listings}
\usepackage{scalerel,xparse}
\usepackage{capt-of,etoolbox}
\newcommand{\dataset}{\mbox{\textsc{{OBJect}}\xspace}}
\newcommand{\datasetnospace}{\mbox{\textsc{{OBJect}}}}
\newcommand{\model}{\mbox{\textsc{{3DIT}}\xspace}}
\newcommand{\modelnospace}{\mbox{\textsc{{3DIT}}}}
\newcommand{\zero}{\mbox{{Zero-1-to-3}}}

\title{\dataset\ \modelnospace: \\Language-guided 3D-aware Image Editing}

\author{
Oscar Michel$^{1}$  \: \: \:
  Anand Bhattad$^{2}$  \: \: \:
  Eli VanderBilt$^{1}$ \\\bf
  Ranjay Krishna$^{1,3}$  \: \: \:
  Aniruddha Kembhavi$^{1}$ \: \: \:
  Tanmay Gupta$^{1}$ \: \: \: \\[0.1in]
  $^1$Allen Institute for Artificial Intelligence, $^2$University of Illinois Urbana-Champaign, \\ $^3$University of Washington 
}

\begin{document}

\makeatletter
\g@addto@macro\@maketitle{
\begin{figure}[H]
    \setlength{\linewidth}{\textwidth}
    \setlength{\hsize}{\textwidth}
    \vspace{-7mm}
    \centering
\includegraphics[width=0.95\textwidth]{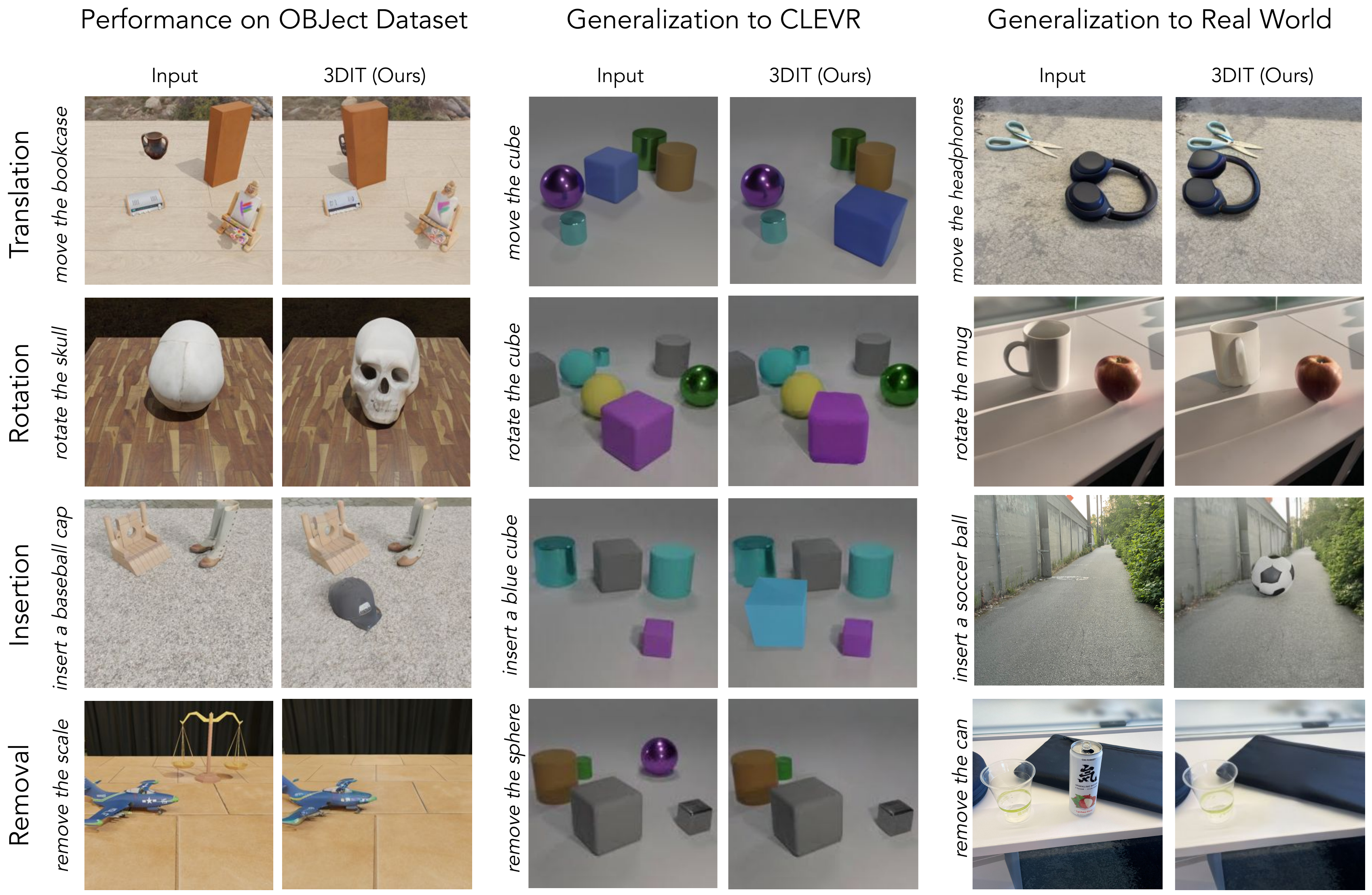}
  \vspace{-6pt}
  \caption{We present \model, a model to edit individual objects in the context of a rich scene with language conditioning. \model\ is able to effectively edit objects while considering their scale and viewpoint, is able to add, remove and edit shadows to be consistent with the scene lighting and is able to account for object occlusions. Training on our new benchmark \dataset, \model\ remarkably generalizes to images in the CLEVR dataset as well as the real world.}

    \label{fig:teaser}
\end{figure}
}
\makeatother  

\maketitle

\begin{abstract}
Existing image editing tools, while powerful, typically disregard the underlying 3D geometry from which the image is projected. As a result, edits made using these tools may become detached from the geometry and lighting conditions that are at the foundation of the image formation process. In this work, we formulate the new task of language-guided 3D-aware editing, where objects in an image should be edited according to a language instruction \textit{in context} of the underlying 3D scene. To promote progress towards this goal, we release \dataset: a dataset consisting of $400$K editing examples created from procedurally generated 3D scenes. Each example consists of an input image, editing instruction in language, and the edited image. We also introduce \model: single and multi-task models for four editing tasks. Our models show impressive abilities to understand the 3D composition of entire scenes, factoring in surrounding objects, surfaces, lighting conditions, shadows, and physically-plausible object configurations. 
Surprisingly, training on only synthetic scenes from \dataset, editing capabilities of \model\ generalize to real-world images. More information can be found on the project page at \url{https://prior.allenai.org/projects/object-edit}.
\end{abstract}

\section{Introduction}
\label{sec:introduction}

In today's visually-oriented society, the art of image editing has become an indispensable necessity. With the proliferation of camera phones and influences from social media platforms, amateur photographers want to transform ordinary snapshots into visual masterpieces. Unfortunately, the process of image editing is still in its infancy. Professional tools such as Photoshop allow pixel-level edits that can adjust lighting, insert objects, remove clutter, and introduce new shadows; however, these tools, with their steep learning curves are often daunting for novices. With the hopes of pulling image editors out from the minutiae of painstaking pixel-level edits, generative models have been heralded as a promise for object-level edits~\cite{ramesh2022hierarchical,ramesh2021zero,ho2020denoising,nichol2021improved}. 

Unfortunately, object-centric editing—translating or rotating an object while preserving the 3D geometry of the original photograph—is out of reach for generative models~\cite{hertz2022prompt, gu2021stylenerf, wang2022score, liu2023zero, tang2023make}.
Although recent strides can take a segmented object and rotate and translate it, they typically operate on objects in isolation and often disregard any scene and lighting context~\cite{liu2023zero,poole2022dreamfusion,tang2023make}.
Others require multiple viewpoints to reconstruct an object in 3D~\cite{chan2022efficient,gu2021stylenerf,wang2022score}.
There is a need for models that can edit objects from a single image while preserving the structure of 3D objects and re-render shadows for the edited scene with the original lighting conditions.

To enable 3D-aware editing of objects in an image, we introduce \datasetnospace, \textbf{Obj}averse \textbf{E}diting in \textbf{C}ontex\textbf{T}, a large-scale benchmark to train and evaluate language-conditioned models that edit objects in images.
We develop \dataset\ by combining Objaverse~\cite{deitke2022objaverse}, a recent 3D asset library, and Blender~\cite{blender}, a 3D rendering engine. 
\dataset\ contains 400k editing examples derived from procedurally generated 3D scenes. Scenes consist of up to four objects, chosen from 59k unique objects, placed on a flat textured surface with an environment lighting map, a three-point lighting system that moves with the camera, and a directional light. 
As shown in  Figure~\ref{fig:teaser}, we support four types of object edits: (a) translation across the surface; (b) rotating around the axis orthogonal to the surface; (c) inserting new objects; and (d) removing existing ones. Our 3D rendering engine ensures that all edits are physically plausible and the generated images capture realistic changes in 3D geometry, illumination, and shading resulting from the underlying edit. For instance, rotation and translation require maintaining contact with the surface; inserting new objects requires identifying stable supported poses for new objects; and removing objects often requires rendering occluded objects.
Each image contains a language instruction describing one of the four edits and a resulting ground truth edited image.
Edited images are evaluated using quantitative metrics that capture realism and faithfulness to the ground truth.

We also introduce \model\ (\textbf{3}D-aware \textbf{D}iffusion \textbf{I}mage-editing with \textbf{T}ext), a model which supports each of the four manipulation tasks with language conditioning. \model\ is initialized with the \zero~\cite{liu2023zero} diffusion model (which was trained to perform novel view synthesis) and finetuned on the \dataset\ dataset for object-centric image editing. The resultant model has effectively been obtained using a three-stage learning curriculum, starting with massive stable diffusion's web-scale pre-training on image-text pairs, followed by \zero's pre-training stage to enhance the model's understanding of 3D objects, and finally with fine-tuning on \dataset\ to enable object-centric edits.

On \dataset's test images, 3DIT outperforms baselines across all four tasks on metrics that capture the faithfulness of the scene edit. Given the known limitations of automatic quantitative metrics, we also provide a human evaluation study, where \model's outputs are preferred to the baselines over $70\%$ of the time. Edits produced by \model\ tend to preserve the original scene's structure and not just the edited object. \model\ preserves the scale and viewpoint of objects, it removes and adds appropriate shadows wherever necessary, and even infills previously occluded portions of the image when the occluder is translated or removed. A multi-task variant of \model\ performs well despite having to support all four transformations using a single set of parameters. 
Finally, \model\ generalizes surprisingly well to new image domains such as CLEVR, a popular synthetic dataset for visual reasoning, as well as real-world images (see Figure~\ref{fig:teaser}). This highlights \model's remarkable capability given that \dataset\ is a synthetic, procedurally generated dataset.

\section{Related work}
\label{sec:related_work}
Much of this work has been inspired by the complexity of editing objects in real-world scenes. It begins to connect old ideas of editing objects in images with today's generative models.

\textbf{Image editing with generative models:} The goals of editing objects and semantic regions in images with language have been active for over a decade~\cite{laput2013pixeltone}.
Back then, productizable edits were limited to simple changes like cropping, colorization and resizing to complex procedures such as object removal, addition, and rearrangement~\cite{reinhard2001color, hertzmann2001image, efros2001image, liao2012subdivision, deshpande2017learning, zhu2017unpaired, gatys2016image, bhattad2020cut, ulyanov2018deep, park2019semantic}.
Traditionally, these tasks were performed manually using tools like Adobe Photoshop. 
However, the origin of Generative Adversarial Networks (GANs)~\cite{goodfellow2014generative} revolutionized the field, propelling significant strides toward automation. StyleGAN~\cite{karras2019style, karras2020analyzing, karras2021alias} notably facilitated intricate modifications to the synthesized images, paving the way for sophisticated GAN-based editing techniques with greater control and flexibility~\cite{shen2020interfacegan, zhu2020indomain, chong2021stylegan, richardson2021encoding, chong2021jojogan, bhattad2023StylitGAN, abdal2021styleflow, shoshan2021gan}. Since then, advancements in generative image architectures have been marked by the emergence of diffusion models~\cite{dhariwal2021diffusion}. When coupled with the availability of large-scale image-text datasets~\cite{schuhmann2022laion}, these models have facilitated the generation of high-fidelity, diverse scenes~\cite{nichol2021glide, ramesh2021zero, rombach2022high, saharia2022photorealistic, kang2023scaling}. Concurrent with these developments, a new wave of image editing methodologies utilizing these large-scale diffusion models have been introduced~\cite {zhang2023adding, mou2023t2i, kawar2023imagic, hertz2022prompt, kumari2022customdiffusion, epstein2023selfguidance}.
Despite these advancements, models lack the 3D awareness necessary for maintaining geometric and lighting consistency. Our dataset, \datasetnospace, aims to bridge this gap by enhancing existing methods and serving to evaluate future methodologies.

\textbf{3D-aware image editing:} A host of recent research, including StyleNeRF~\cite{gu2021stylenerf}, EG3D~\cite{chan2022efficient}, SJC~\cite{wang2022score}, DreamFusion~\cite{poole2022dreamfusion}, Zero-1-to-3~\cite{liu2023zero}, and Make-It-3D~\cite{tang2023make}, has explored lifting 2D images to 3D. By contrast, our model—\modelnospace— comprehensively considers the entire scene, not just the object of interest, encompassing geometry, lighting, and other salient attributes of the background.

\textbf{Scene rearrangement:} Current research in scene rearrangement tasks primarily involve solving rearrangement from robotic manipulation and embodied agents~\cite{king2016rearrangement, qureshi2021nerp, radford2021learning, liu2022structformer} to provide more intuitive and human-like commands for scene manipulation and navigation. Specific attempts have also been made to apply these techniques to room rearrangements~\cite{wang2020scenem, weihs2021visual, wei2023lego} using datasets like AI2-THOR~\cite{kolve2017ai2}, Habitat~\cite{szot2021habitat}, Gibson~\cite{xiazamirhe2018gibsonenv} or 3D-FRONT~\cite{fu20213d}. For instance, LegoNet~\cite{wei2023lego} focuses on room rearrangements without the need to specify the goal state, learning arrangements that satisfy human criteria from professionally arranged datasets provided by 3D-FRONT~\cite{fu20213d}.
Distinct from these works, our research introduces a unique perspective. We focus on object-level rearrangements with a primary emphasis on 3D-aware image editing using language instructions. \model\ is trained with \dataset\ to edit scenes with a high degree of realism and 3D coherence.

\textbf{3D asset datasets:} A diverse set of 3D asset dataset such as ShapeNet~\cite{chang2015shapenet} and the recent Objaverse~\cite{deitke2022objaverse} have played a pivotal role in 3D computer vision. ShapeNet provides a richly-annotated, large-scale dataset of 3D shapes that has found numerous applications in object recognition, scene understanding, and 3D reconstruction. Objaverse has offered a large collection of 3D objects that are semantically segmented and paired with natural language descriptions. Objaverse has been instrumental in the construction of \dataset\ and also advancing several other related research areas, including generating textured meshes~\cite{chen2023text2tex, gao2022get3d, gupta20233dgen} zero-shot single image 3D generation~\cite{liu2023zero} and enriching simulators~\cite{kolve2017ai2thor,deitke2022procthor} for Embodied AI.

\textbf{Synthetic datasets for vision models:} Diagnostic datasets such as CLEVR~\cite{johnson2017clevr} and CLEVERER~\cite{yi2019clevrer} provide a rigorous test bed for the visual reasoning abilities of models. They contain synthetically generated images of 3D scenes with simple primitives and associated questions that require an understanding of the scene's objects, attributes, and relations to answer correctly. Kubric~\cite{greff2022kubric} is an image and video dataset generation engine that can model physical interactions between objects.  In a similar vein, \dataset\ offers procedurally generated scenes of commonly occurring natural objects derived from ObjaVerse~\cite{deitke2022objaverse} with configurable 3D objects and associated language instructions.

\textbf{Benchmarks for image editing:} There is currently a scarcity of benchmarks to evaluate generative models~\cite{hu2023tifa}, especially for 3D scene editing. Existing ones, including light probes~\cite{wang2020people}, repopulating street scenes~\cite{wang2021repopulating}, GeoSim~\cite{chen2021geosim} and CADSim~\cite{wangcadsim} are not publicly available. Our presented \dataset\ benchmark will be made publicly available.

\section{\dataset: A benchmark for \textbf{Obj}ect \textbf{E}diting in \textbf{C}ontex\textbf{t}}

Our goal is to design and evaluate image editing models capable of editing objects in scenes. To enable training and evaluation of such models, we develop \dataset. \dataset\ contains scenes with multiple objects placed on a flat textured surface and illuminated with realistic lighting.  These edits are described to the model using a combination of language and numerical values (e.g. pixel coordinates and object rotation angle). All edits result in structural changes to the scene which in turn affect illumination changes such as inter-object reflections and shadows. The model does not have access to the underlying 3D scene (including object segmentations, locations, 3D structure, and lighting direction); it must infer these from the input pixels.  

\subsection{Object editing tasks}
\dataset\ supports four fundamental object editing tasks: Each of the following manipulations targets a single object within a scene that may contain multiple objects. We now describe each task and the capabilities required from an image editing model to succeed at the task. For specifying locations in an image, we use a coordinate system where (0,0) represents the bottom-left corner and (1,1) the top-right corner. Objects are specified  in each task using their crowdsourced descriptions.

\noindent\textbf{Translation}: Given the x-y coordinates of a target location, a specified object is moved from its original location in the scene to the target location while preserving its angular pose and surface contact. Since the camera is fixed relative to the scene, a change in object location requires to model to synthesize newly visible portions of the object. The model is required to change the object's scale in the image due to perspective projection i.e. the objects should appear smaller when moved further away from the camera and vice-versa. The new location may also result in drastically different illumination of the object.

\noindent\textbf{Rotation}: A specified object is rotated counter-clockwise around the vertical axis passing through the object's center of mass and perpendicular to the ground by a given angle. To succeed, the model must localize the object, extrapolate the object's shape from a single viewpoint, and re-imagine the scene with the rotated object. Rotating objects leads to intricate changes to the shadow projected on the ground plane which are challenging to accurately produce.

\begin{wrapfigure}{r}{0.5\textwidth}
  \vspace{-0.3in}
  \begin{center}
    \includegraphics[width=0.4\textwidth]{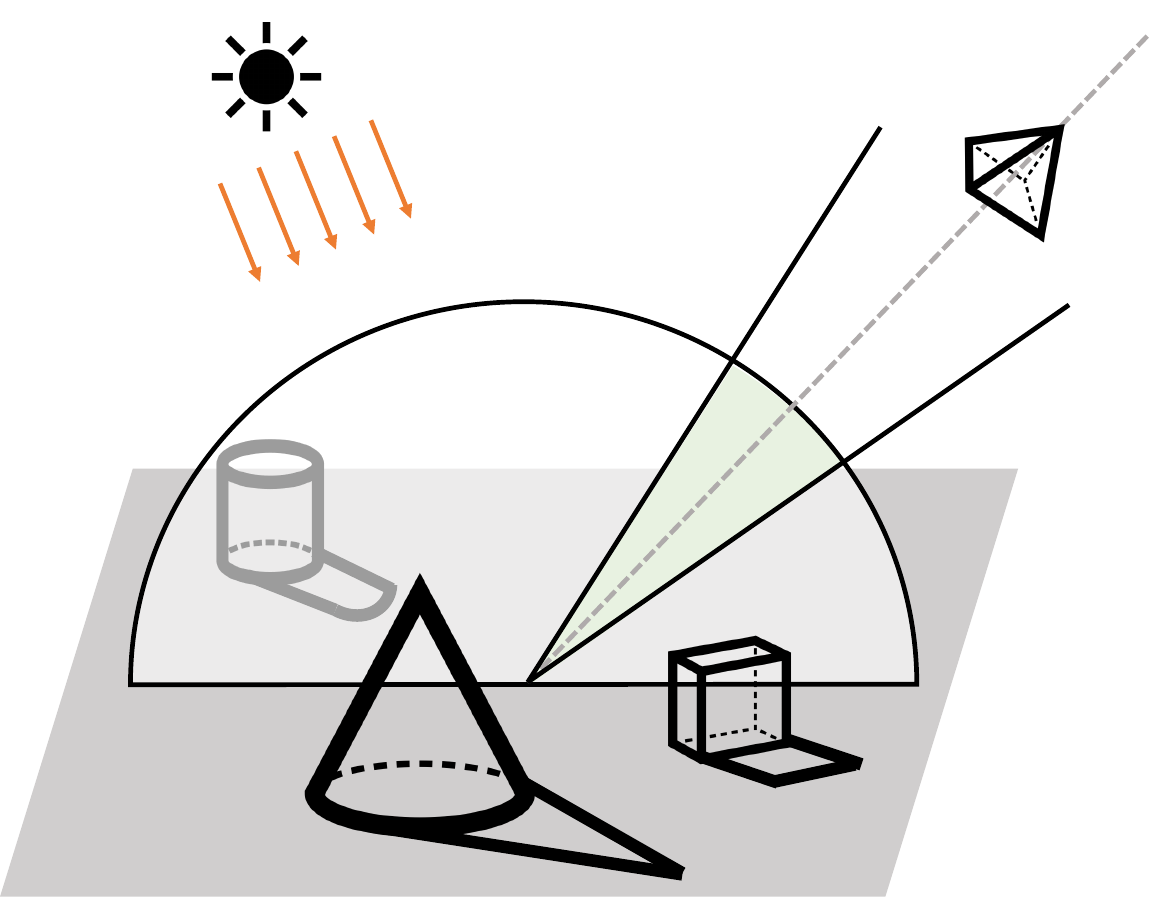}
  \end{center}
  \caption{Scene generation in \dataset\ depicting camera constraints, directional lighting (environment and three-point lighting not shown), and the resulting object shadows.}
  \vspace{-0.5in}
\end{wrapfigure}

\noindent\textbf{Insertion}: Given a language description, an object matching the description is added to the scene at a designated x-y location. The model must perform object generation at the desired location with stable pose and surface contact. Besides modeling the object shape, the model also needs to understand the interaction of the geometry with scene lighting to generate a realistic shadow for the object.

\noindent\textbf{Removal}: A specified object is removed from the scene. The model must not only be able to locate and segment the object, but also in-paint the object region using scene context. This often requires inpainting an object that was previously partially or fully occluded. 

\subsection{Benchmark curation}
Paired image-\&-text data is plentiful on the internet and large corpora are commonly used to train text-to-image models. However, there is a lack of image editing data consisting of initial and edited image pairs, with a description of the edit. Gathering such a dataset at scale from the real world requires significant manipulation and annotation effort. Our key insight is that while object manipulation data is difficult to acquire, it is much easier to synthesize large volumes of this data leveraging the latest advances in photorealistic rendering and large 3D asset libraries. Therefore, \dataset\ contains procedurally generated 3D scenes rendered with objects from these asset libraries.

\noindent\textbf{Object source}. \dataset\ scenes are constructed using one to four 3D objects from the Objaverse dataset\cite{deitke2022objaverse}. The entire Objaverse dataset contains more than $800$k assets. Since objaverse contains objects with errors (some objects are not fully rendered or contain no texture), we filter the objects down to a set of $59$k via a combination of Sketchfab metadata-based filtering and crowdsourcing. The resulting objects all have various textures, are easily recognizable, are of high quality and resolution, are free of copyrighted material, are in isolation (as opposed to a single asset with multiple objects), and are free floating (so that they may be placed on any surface in the generated scenes).

Each of these assets is annotated with one of $1613$ unique semantic categories using crowdsourcing. Workers were shown a rotating 3D rendering of a particular object and asked to apply a category label; they were provided with a handy autocomplete list of roughly $1400$ categories sourced from LVIS~\cite{Gupta2019LVISAD}categories. If, however, workers were unable to find an appropriate category, they had the option of generating a new category. After this they were asked to write a sentence that describes the object, pointing out any interesting or noteworthy details that would distinguish it from other objects in the same category. Finally, category names were cleaned up to remove spelling errors; we removed unusual or rare categories.

We randomly choose $1513$ categories to be seen during training while holding out the remaining $100$ as unseen categories for validation and testing. This category split helps quantify the generalization gap in editing previously seen vs novel objects. We use a library of $17$ texture maps obtained from \cite{opengameart} to simulate wooden, cobblestone, and brick flooring for the scenes.

\noindent\textbf{Scene construction}. To have the scene layout and lighting appear natural, we specify a set of constraints and perform rejection sampling for selecting physically plausible object instances, scene layout, camera positions, and lighting parameters. We limit all scenes to a minimum of one and a maximum of four objects. To identify a natural resting pose for these objects, we perform a physical simulation in Blender where we drop each object onto an XY ground plane and record its resting pose. Then to identify object placements, we sample a bounding box of the same x-y aspect ratio as the object and uniformly scale the object to lie in this bounding box. We ensure that objects, when rotated, do not intersect each other: bounding boxes who's circumscribed circles intersect are rejected. To avoid tiny objects being placed in the same scene as very large objects, we enforce the ratio between the smallest and longest largest side of each bounding box to be greater than $0.8$. We randomly place the camera in the upper hemisphere surrounding the plane and point it towards the origin which lies on the ground plane. We further constrain the camera elevation angle from the ground between $40^{\circ}$ to $80^{\circ}$ to ensure that the viewing angle is neither too close to the ground nor completely vertical which are both relatively unnatural. In each scene, there is a designated object that is manipulated. If this object is not visible from the camera, we move the camera away from the origin until the object is visible both before and after the manipulation. 

\noindent\textbf{Scene lighting}. We use several light sources to realistically illuminate the scene. First, we add a random environment lighting map, which are special images that capture the light in a real-world scene from all directions, giving the impression that our constructed scenes are imbedded in various indoor and outdoor locations in the real world. We download 18 of these environment maps with CC0 licences from \url{https://polyhaven.com/}. Next, we add a three-point lighting system that automatically adapts to the camera view. This involves placing the key light for primary illumination, the fill light to soften key light shadows, and the back light to distinguish the subject from the background. These lights serve to effectively shade the objects in the front of the camera so that their 3D form is apparent.  Finally, the scene is illuminated with directional lighting with the direction randomly sampled within a conical neighborhood around the negative-z direction to simulate an overhead light source. This consists of parallel rays emitted by a single light source infinitely far away and therefore can be specified by intensity and direction without specifying a source position. %

We generate 100k training examples for each task, and 1024 scenes for validation and testing. The 3D scenes are automatically generated using Blender and its Cycles ray tracer for rendering each scene. We also render segmentation masks that denote object instances, plane and background pixels for all scenes.

\label{sec:benchmark}
\section{\model: a scene-aware editing model}
\label{sec:model}

\noindent\textbf{Task setup.} Consider a 3D scene, $S$, filled with multiple objects. Let $x_1 \in \mathbb{R}^{H\times W\times 3}$ represent an image of this scene produced by a rendering function $f$. Let $l$ represent the text description of the edit, and $v$ represent the task-specific numerial values (i.e. angle for the rotation task and x,y coordinates for removal, insertion, and translation) to describe the desired edit to the scene $S$. In this paper, we consider object-centric manipulations including rotating, translating, inserting, and removing objects. Manipulating the objects in $S$ can yield a new image $x_2 = f(M(S, l, v))$, where $M$ applied the transformation $l, v$ in 3D.

Our goal is to produce $x_2$ without access to the 3D scene $S$ and instead, directly editing the source image $x_1$. Importantly, we have no explicit information about the scene (including scene geometry and layout), no explicit information about the lighting (such as its location and intensity), and no access to the camera parameters. All this information must be implicitly inferred from the single source image $x_1$. Concretely, we wish to produce the target image $x_2 = \hat{f}_\theta(x_1, l, v)$, where $\hat{f}$ is a learned function with parameters $\theta$.

\noindent\textbf{Background.} Diffusion models~\cite{rombach2021highresolution} have recently shown spectacular results in generating images conditioned on text descriptions. These models consist of an encoder $\mathcal{E}$ that maps an image $x$ into a latent code $z=\mathcal{E}(x)$, a decoder, $\mathcal{D}$ that can map a latent code back to image space, and a U-Net $\epsilon_{\theta}$ with learned parameters $\theta$ used for denoising. Some diffusion models are trained on large training corpora such as LAION-5B~\cite{schuhmann2022laion5b} and are able to produce high-quality high-resolution images that faithfully represent input text descriptions. The recently proposed \zero\ model\cite{liu2023zero} finetunes image-conditioned Stable Diffusion\cite{imcondsd} on the task of generating an image of a single object from a novel viewpoint, conditioned on an input view and a relative camera transformation. 

\noindent\textbf{\model.} Our model, \model, builds upon \zero. We design $\hat{f}_\theta(\cdot)$ using the same base architecture but make changes to its conditioning module $c_\theta(\cdot)$. Our changes enable the conditioning module to accept edit instructions in the form of language and location information to precisely define the desired edit. In the cross-attention conditional module, \zero\ uses a CLIP image encoding to represent the initial image, followed by concatenating a four-dimensional vector encoding camera pose information. This $772$-dimensional vector gets passed through a multi-layered perceptron (MLP) to map it back down to a size of $768$ dimensions. Similarly, we encode the source image $x_1$ using the same CLIP image encoder. We encode $v$ and concatenated the vector with the image representation and feed it into the MLP. Next, we append the MLP outputs with edit text tokens $l$, which are extracted using CLIP's text encoder.

We finetune our model from the $16,500$-step checkpoint of \zero. During training, the network takes a noised latent encoding of $z_t$, timestep $t$ and conditioning information $c(x_1, l, v)$, where $z_t$ is the latent representation of the target image at time step $t$.
and produces a denoising score estimate $\epsilon_\theta(z_t, t, c(x_1, l, v))$ 
where $c(\cdot) \in\mathbb{R}^{768\times N}$ outputs a sequence of conditional embedding vectors. 
We finetune the network with the standard diffusion loss~\cite{ho2020denoising,rombach2022high}:
\[\min_\theta \mathbb{E}_{z \sim \mathcal{E}_\theta(x_1), t, \epsilon \sim \mathcal{N}(0, 1)} || \epsilon - \epsilon_{\theta}(z_t, t, c(x_1, l, v)) || \]

The entire training procedure for \model\ resembles a three-stage curriculum. The first pre-training stage with billions of LAION image-text pairs teaches the model to produce rich imagery for text descriptions constructed from a very large vocabulary. These images contain diverse backgrounds and a variety of objects within them. Although the model can faithfully produce imagery, its ability to perform fine-grained manipulations of objects remains limited. The second pre-training stage with from \zero's millions of image-image pairs of different viewpoints enhances the model's understanding of 3D object shape and camera projection. The third training stage, outlined in this paper, uses \dataset\ and teaches the model to manipulate objects in physically plausible scenes. We show that the resulting model can now edit individual objects without any 3D explicit information.

\section{Experiments}
\label{sec:experiments}

We now present experiments to evaluate our \model\ model. First, we evaluate single task variants of \model, i.e. one model for each of the four tasks -- object rotation, translation, insertion and removal. For each of these tasks, we evaluate the performance of the model on novel scenes with objects seen at training time, and with objects unseen at training time. We also provide evaluations for a multi-task model -- trained to perform all four tasks.

\begin{figure}
  \centering
  \includegraphics[width=\textwidth]{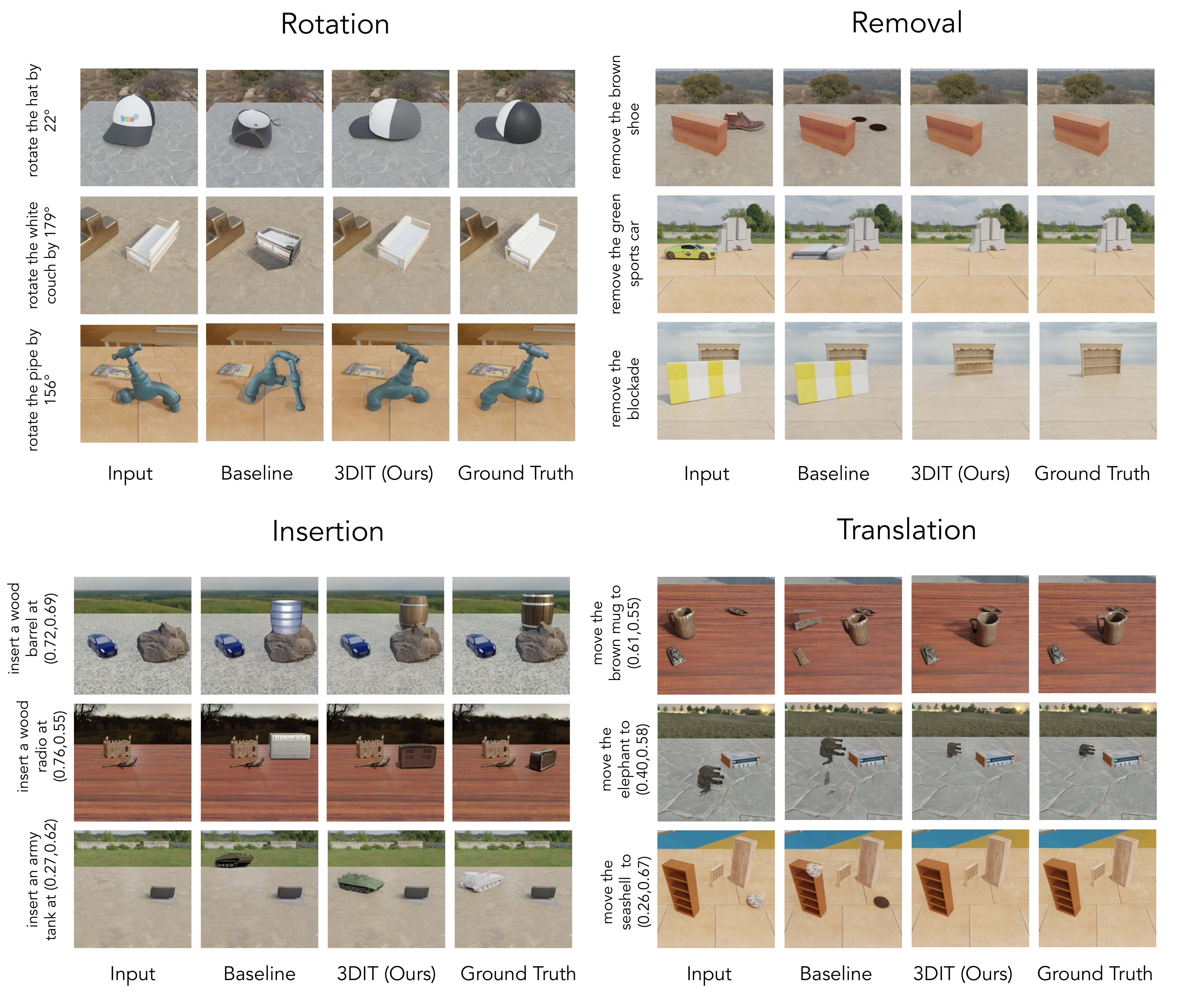}
  \label{fig:single_task}
  \vspace{-1em}
  \caption{Generated examples from \model\ as well as baselines for each of the four tasks in the \dataset\ benchmark.}
  \vspace{-1em}
\end{figure}
\subsection{Baselines}

For each of the four tasks, we create strong baselines inspired by recent approaches like VisProg~\cite{Gupta2022VisualPC} and Socratic models~\cite{Zeng2022SocraticMC} that chain multiple foundation models together to create performant systems for various tasks including image editing. 

\noindent \textbf{Removal:} Removing an object from a scene requires segmenting out the object and inpainting the region. Since a point within the object region is provided as input, we use SAM~\cite{Kirillov2023SegmentA} to generate a segmentation mask. We first use SAM in the generation mode to get candidate masks for the entire scene and select the mask which contains the point and occupies no more than a third of the area of the entire image. If no such mask is found, we attempt to get a mask by directly using the point as input to SAM to get a mask. Then, we use Stable Diffusion (SD) to inpaint the masked region using the prompt ``a rendering of an uncluttered textured floor with no objects''. We found that using the selected region-encompassing bounding box as the mask works better than using the fine-grained segmentation mask. 

\noindent \textbf{Insertion:} This baseline uses SD and the target location to re-imagine the scene with an object of the provided category. The final image is generated by SD using the prompt ``a 3D rendering of {category} on a textured floor" conditioned on the initial image and a fixed-size ($200\times200$) square mask around the target location.

\noindent \textbf{Translation:} Translation requires localizing the object given the category name, removing it from the initial location, and inserting it at the target location. We use OWL-ViT~\cite{Minderer2022SimpleOO} to localize the object given the category name. The detected bounding box is fed into SAM to generate the object segmentation mask which is then used for inpainting similar to the Removal baseline. Finally, the segmented object is composited at the target location.  

\noindent \textbf{Rotation:} Here we use \zero~\cite{liu2023zero} as a baseline which requires the object to be tightly centered in the image with a white background. So, we first localize the object using OWL-ViT, crop the localized region, and segment it using SAM to create the appropriate input for \zero\ for performing the rotation. The rotated object is composited back onto the image and the remaining unfilled regions are inpainted using SD.

\subsection{Quantitative evaluation}
\begin{table}[ht]
\centering
\small
\caption{Quantitative evaluation using generated samples. For each method, four samples per test image were generated. The best image according to the PSNR metric is selected to represent each sample, and these values are averaged across samples. To ensure that the metrics focus on the transformed object and not the background which mostly remains unchanged, metrics are computed using the region around the transformed object's mask.}
\label{tab:quant_eval}
\begin{tabular}{lcccccccc}
\toprule
& \multicolumn{4}{c}{Seen Objects} & \multicolumn{4}{c}{Unseen Objects} \\
Model & PSNR $\uparrow$ & SSIM $\uparrow$ & LPIP $\downarrow$ & FID $\downarrow$ & PSNR $\uparrow$ & SSIM $\uparrow$ & LPIP $\downarrow$ & FID $\downarrow$\\
\midrule
\multicolumn{9}{c}{\textit{Task: Translation}} \\
\midrule
Baseline & 13.699 & \textbf{0.309} & 0.485 & 0.942 & 14.126 & \textbf{0.326} & \textbf{0.467} & 0.968 \\
\model (1-task) & 14.546 & 0.273 & 0.494 & 0.254 & 14.400 & 0.262 & 0.498 & 0.261 \\
\model (Multitask) & \textbf{15.21} & 0.300 & \textbf{0.472} & \textbf{0.244} & \textbf{15.200} & 0.292 & 0.477 & \textbf{0.253} \\

\midrule
\multicolumn{9}{c}{\textit{Task: Rotation}} \\
\midrule
Baseline & 13.179 & 0.269 & 0.540 & 0.997 & 12.848 & 0.270 & 0.538 & 1.693 \\
\model (1-task) & 16.828 & \textbf{0.386} & \textbf{0.428} & 0.291 & \textbf{16.293} & \textbf{0.372} & \textbf{0.445} & 0.280 \\
\model (Multitask) & \textbf{16.859} & 0.382 & 0.429 & \textbf{0.248} & 16.279 & 0.366 & 0.447 & \textbf{0.236} \\
\midrule
\multicolumn{9}{c}{\textit{Task: Insertion}} \\
\midrule
Baseline & 12.297 & \textbf{0.269} & 0.594 & 0.969 & 12.542 & \textbf{0.275} & 0.584 & 1.325 \\
\model (1-task) & 13.469 & 0.267 & \textbf{0.549} & 0.254 & 12.974 & 0.261 & \textbf{0.566} & 0.233 \\
\model (Multitask) & \textbf{13.630} & 0.263 & 0.551 & \textbf{0.222} & \textbf{13.088} & 0.261 & 0.568 & \textbf{0.214} \\
\midrule
\multicolumn{9}{c}{\textit{Task: Removal}} \\
\midrule
Baseline & 12.494 & 0.383 & 0.465 & 0.801 & 12.123 & 0.379 & 0.459 & 1.047 \\
\model (1-task) & 24.937 & \textbf{0.588} & 0.254 & 0.241 & 24.474 & 0.561 & \textbf{0.260} & 0.258 \\
\model (Multitask) & \textbf{24.980} & 0.585 & \textbf{0.249} & \textbf{0.236} & \textbf{24.661} & \textbf{0.568} & \textbf{0.260} & \textbf{0.240} \\
\bottomrule
\end{tabular}
\end{table}

We follow Zero-1-to-3 and use four metrics to automatically evaluate the quality and accuracy of the edited image - PSNR, SSIM, LPIPS, and FID. The first 3 directly compare the prediction to the ground truth image, while FID measures the similarity between the predicted and ground truth sets of images. Instead of computing the metrics for the whole image, we focus on the region where the edits are targeted. To do this, we simply use the ground truth segmentation mask to crop the targeted rectangular region of interest prior to computing the metrics. Since our model, as well as our baselines, can generate multiple solutions for each input, our evaluation considers the best-of-four prediction as per the SSIM metric to compute the final scores for all metrics. This considers the typical use case for editing applications where a user has the flexibility to pick from a range of generated solutions. We report metrics separately for seen and unseen object categories.

\begin{figure}
  \centering
  \includegraphics[height=0.75\textwidth]{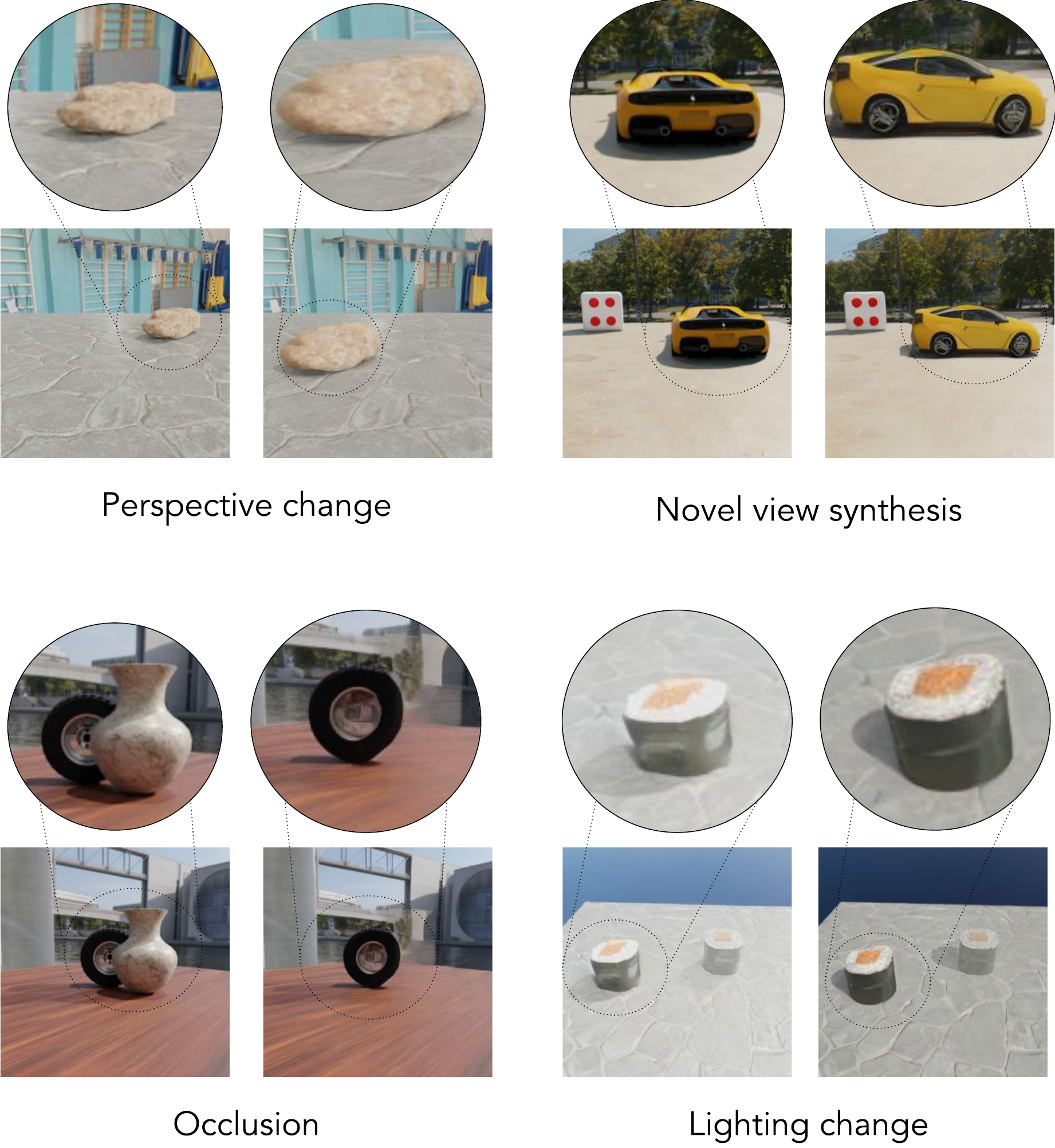}
  \vspace{1em}
  \caption{The figure shows the ability of \model\ to handle various challenges of 3D-aware image editing such as: (a) (\textit{Top left}) perspective size changes; (b) (\textit{Top right}) synthesizing novel view points; (c) (\textit{Bottom left}) generating occluded regions; (d) (\textit{Bottom right}) accounting for scene lighting while rendering objects and their shadows.} %
  \vspace{-1em}
  \label{fig:challenges}
\end{figure}

Table~\ref{tab:quant_eval} presents quantitative evaluations for \model\ in comparison to the baselines. \model outperforms the baselines for all four tasks at the metrics PSNR, SSIM and LPIP. Notably, the multi task model does well in comparison to the single task variant, in spite of having to learn 4 tasks using the same number of learnable parameters. The FID scores for the baseline models tend to be higher. This is because the baselines tend to cut/paste objects in the image (for e.g. in the translation task), which retains image fidelity, even if the scale of the object is incorrect. \model\ on the other hand does not explicitly cut/paste segments and instead must render them using the diffusion process, and is thus prone to a poorer fidelity. On the contrary, our model is able to properly account for a variety of challenging changes to the underlying 3D scene when editing images, as shown in Figure ~\ref{fig:challenges}. Its worth noting that the automatic evaluation metrics have limitations and often do not capture editing nuances encompassing geometry, lighting, and fidelity to the instruction. This motivates the need for human evaluation studies.

\subsection{Human evaluation studies}

We conduct human preference evaluations between \model\ and the relevent baseline by showing two images and asking annotators to select the one that best matches the ground truth image. We measure 
(1) \textbf{Geometric consistency} -- This requires humans to consider the geometric correctness of the transformed object, including the scale, positioning of the object on the ground plane and its relationship to other objects. It also requires humans to consider the correctness of other objects in the scene which may get occluded or unoccluded as a result of the transformation.  
the source caption. 
(2) \textbf{Lighting consistency} -- This requires humans to consider the lighting correctness of the transformed object, including the direction and scale of the shadow as a result of the directional lighting. It also requires humans to consider the correctness of the shadows of other objects in the scene which may get occluded or unoccluded as a result of the transformation.  
Both evaluations also allow a third option (Tie) to be selected. Each pairwise evaluation is carried out for 30 test samples.

Table~\ref{tab:human_eval} presents a human evaluation study of the \model\ model (in a single task setting) in comparison to the corresponding baseline for all four tasks. \model\ is heavily favored by humans, consistently obtaining preference scores of 70 \% and more across all four tasks for geometric as well lighting consistency. The tied scores refer to instances where both models did exceedingly poorly and where both models did a close to perfect job.

For the translation task, \model\ is able to scale the object  appropriately, as well rendering the shadow correctly. The baseline, in particular, does a poor job of the shadow and gets the scale wrong, leading to a physically implusible image. For the rotation task, \model\ performs a rotation consistent with the ground plane and also renders a superior shadow. For the removal task, \model\ tends to inpaint occluded objects well, and correctly adjusts their shadows. It also does well at removing the entire extent of the correct object in contrast to the baseline. 

\begin{table}[ht]
\centering
\caption{Outcome of the human evaluation. The table illustrates the evaluators' preferences for \model\, assessed on geometric accuracy and 3D lighting consistency. Baseline methods rarely gained preference due to their limited capacity to maintain geometric quality and lighting consistency.}
\label{tab:human_eval}
\begin{tabular}{lcccccc}
\toprule
Task & \multicolumn{3}{c}{Geometric consistency} & \multicolumn{3}{c}{Lighting consistency} \\
\midrule
 & Baseline & \model \textit{(Ours)} & Tie & Baseline & \model \textit{(Ours)} & Tie \\
\midrule

Translation & 20.0 \% & 73.3 \% & 6.6 \% & 3.3 \% & 80.0 \% & 16.6 \% \\
Rotation & 3.3 \% & 80.0 \% & 16.6 \% & 6.6 \% & 73.3 \% & 20.0 \% \\
Insertion & 13.3 \% & 70.0 \% & 16.6 \% & 10.0 \% & 73.3 \% & 16.6 \% \\
Removal & 3.3 \% & 86.6 \% & 10.0 \% & 0.0 \% & 86.6 \% & 13.3 \% \\
\bottomrule
\end{tabular}
\vspace{-0.1in}
\end{table}

\subsection{Real-world transfer}
\begin{figure}
  \centering
  \includegraphics[width=\textwidth]{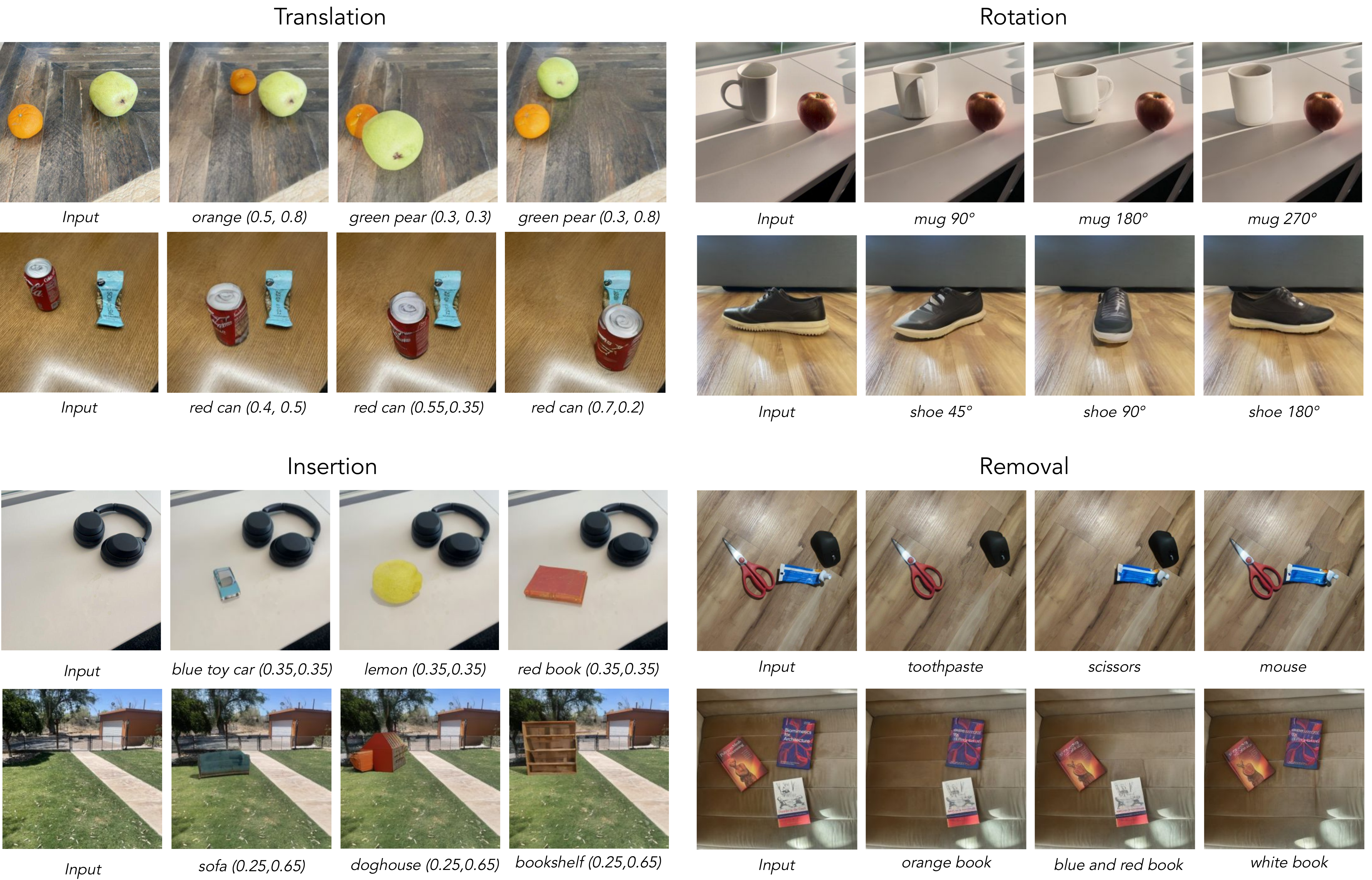}
  \vspace{-1em}
  \caption{\model\ is able to generalize to the real world while only being trained on a synthetic dataset. Here we show varying prompts for each of the four editing tasks.}
  \vspace{-1em}
  \label{fig:real_examples}
\end{figure}
While we train our models on simulated data, we test the model's ability to transfer to real-world images qualitatively. Figure~\ref{fig:real_examples} shows our model's output for different prompts for the same input image for all four tasks. We find these preliminary results encouraging as the outputs not only respect the task description but also look reasonably photo-realistic with appropriate shadows despite never seeing real-world editing examples during training.

\section{Limitations and Broader Impact}

Our work explores the use of synthetic data for training physically plausible and scene-aware image editing models. Given that even training on scenes with limited realism and complexity results in models that transfer well to the real world, there is tremendous potential to significantly improve performance by using more advanced photo-realistic simulators. Finetuning on a small set of hand-crafted real-world editing examples may also improve transfer to real-world images and enable compelling editing applications. Our work leads the way towards easy-to-use and increasingly powerful image editing capabilities for the broader society in the near future. Like any generative model, our work could also potentially be misused for propagating misinformation.

\section{Conclusion}

This work presents \model, a model capable of editing individual objects within images, given a language instruction. \model\ is trained on a  new dataset, \dataset, consisting of 400k 3D scenes procedurally generated using Objaverse objects. \model\ performs well across on \dataset\ and shows promising generalization to CLEVR as well as the real world.

\bibliography{references, ref_anand}

\begin{thebibliography}{10}

\bibitem{abdal2021styleflow}
R.~Abdal, P.~Zhu, N.~J. Mitra, and P.~Wonka.
\newblock Styleflow: Attribute-conditioned exploration of stylegan-generated
  images using conditional continuous normalizing flows.
\newblock {\em ACM Transactions on Graphics (ToG)}, 40(3):1--21, 2021.

\bibitem{opengameart}
O.~G. Art.
\newblock {OpenGameArt.org}.
\newblock \url{https://opengameart.org/}, 2023.
\newblock [Online; accessed 16-May-2023].

\bibitem{bhattad2020cut}
A.~Bhattad and D.~Forsyth.
\newblock Cut-and-paste object insertion by enabling deep image prior for
  reshading.
\newblock In {\em 2022 International Conference on 3D Vision (3DV)}. IEEE,
  2022.

\bibitem{bhattad2023StylitGAN}
A.~Bhattad and D.~Forsyth.
\newblock Stylitgan: Prompting stylegan to generate new illumination
  conditions.
\newblock In {\em arXiv}, 2023.

\bibitem{chan2022efficient}
E.~R. Chan, C.~Z. Lin, M.~A. Chan, K.~Nagano, B.~Pan, S.~De~Mello, O.~Gallo,
  L.~J. Guibas, J.~Tremblay, S.~Khamis, et~al.
\newblock Efficient geometry-aware 3d generative adversarial networks.
\newblock In {\em Proceedings of the IEEE/CVF Conference on Computer Vision and
  Pattern Recognition}, pages 16123--16133, 2022.

\bibitem{chang2015shapenet}
A.~X. Chang, T.~Funkhouser, L.~Guibas, P.~Hanrahan, Q.~Huang, Z.~Li,
  S.~Savarese, M.~Savva, S.~Song, H.~Su, et~al.
\newblock Shapenet: An information-rich 3d model repository.
\newblock {\em arXiv preprint arXiv:1512.03012}, 2015.

\bibitem{chen2023text2tex}
D.~Z. Chen, Y.~Siddiqui, H.-Y. Lee, S.~Tulyakov, and M.~Nie{\ss}ner.
\newblock Text2tex: Text-driven texture synthesis via diffusion models.
\newblock {\em arXiv preprint arXiv:2303.11396}, 2023.

\bibitem{chen2021geosim}
Y.~Chen, F.~Rong, S.~Duggal, S.~Wang, X.~Yan, S.~Manivasagam, S.~Xue, E.~Yumer,
  and R.~Urtasun.
\newblock Geosim: Realistic video simulation via geometry-aware composition for
  self-driving.
\newblock In {\em Proceedings of the IEEE/CVF Conference on Computer Vision and
  Pattern Recognition}, 2021.

\bibitem{chong2021jojogan}
M.~J. Chong and D.~Forsyth.
\newblock Jojogan: One shot face stylization.
\newblock {\em arXiv preprint arXiv:2112.11641}, 2021.

\bibitem{chong2021stylegan}
M.~J. Chong, H.-Y. Lee, and D.~Forsyth.
\newblock Stylegan of all trades: Image manipulation with only pretrained
  stylegan.
\newblock {\em arXiv preprint arXiv:2111.01619}, 2021.

\bibitem{blender}
B.~O. Community.
\newblock Blender - a 3d modelling and rendering package, 2018.

\bibitem{deitke2022objaverse}
M.~Deitke, D.~Schwenk, J.~Salvador, L.~Weihs, O.~Michel, E.~VanderBilt,
  L.~Schmidt, K.~Ehsani, A.~Kembhavi, and A.~Farhadi.
\newblock Objaverse: A universe of annotated 3d objects.
\newblock {\em arXiv preprint arXiv:2212.08051}, 2022.

\bibitem{deitke2022procthor}
M.~Deitke, E.~VanderBilt, A.~Herrasti, L.~Weihs, J.~Salvador, K.~Ehsani,
  W.~Han, E.~Kolve, A.~Farhadi, A.~Kembhavi, et~al.
\newblock Procthor: Large-scale embodied ai using procedural generation.
\newblock {\em Conference on Neural Information Processing Systems}, 2022.

\bibitem{deshpande2017learning}
A.~Deshpande, J.~Lu, M.-C. Yeh, M.~Jin~Chong, and D.~Forsyth.
\newblock Learning diverse image colorization.
\newblock In {\em Proceedings of the IEEE Conference on Computer Vision and
  Pattern Recognition}, pages 6837--6845, 2017.

\bibitem{dhariwal2021diffusion}
P.~Dhariwal and A.~Nichol.
\newblock Diffusion models beat gans on image synthesis.
\newblock {\em Advances in Neural Information Processing Systems},
  34:8780--8794, 2021.

\bibitem{efros2001image}
A.~A. Efros and W.~T. Freeman.
\newblock Image quilting for texture synthesis and transfer.
\newblock In {\em Proceedings of the 28th annual conference on Computer
  graphics and interactive techniques}, pages 341--346, 2001.

\bibitem{epstein2023selfguidance}
D.~Epstein, A.~Jabri, B.~Poole, A.~A. Efros, and A.~Holynski.
\newblock Diffusion self-guidance for controllable image generation.
\newblock {\em arXiv preprint arXiv:2306.00986}, 2023.

\bibitem{fu20213d}
H.~Fu, B.~Cai, L.~Gao, L.-X. Zhang, J.~Wang, C.~Li, Q.~Zeng, C.~Sun, R.~Jia,
  B.~Zhao, et~al.
\newblock 3d-front: 3d furnished rooms with layouts and semantics.
\newblock In {\em Proceedings of the IEEE/CVF International Conference on
  Computer Vision}, pages 10933--10942, 2021.

\bibitem{gao2022get3d}
J.~Gao, T.~Shen, Z.~Wang, W.~Chen, K.~Yin, D.~Li, O.~Litany, Z.~Gojcic, and
  S.~Fidler.
\newblock Get3d: A generative model of high quality 3d textured shapes learned
  from images.
\newblock In {\em Advances In Neural Information Processing Systems}, 2022.

\bibitem{gatys2016image}
L.~A. Gatys, A.~S. Ecker, and M.~Bethge.
\newblock Image style transfer using convolutional neural networks.
\newblock In {\em Proceedings of the IEEE conference on computer vision and
  pattern recognition}, pages 2414--2423, 2016.

\bibitem{goodfellow2014generative}
I.~J. Goodfellow, J.~Pouget-Abadie, M.~Mirza, B.~Xu, D.~Warde-Farley, S.~Ozair,
  A.~Courville, and Y.~Bengio.
\newblock Generative adversarial networks.
\newblock {\em arXiv preprint arXiv:1406.2661}, 2014.

\bibitem{greff2022kubric}
K.~Greff, F.~Belletti, L.~Beyer, C.~Doersch, Y.~Du, D.~Duckworth, D.~J. Fleet,
  D.~Gnanapragasam, F.~Golemo, C.~Herrmann, et~al.
\newblock Kubric: A scalable dataset generator.
\newblock In {\em Proceedings of the IEEE/CVF Conference on Computer Vision and
  Pattern Recognition}, pages 3749--3761, 2022.

\bibitem{gu2021stylenerf}
J.~Gu, L.~Liu, P.~Wang, and C.~Theobalt.
\newblock Stylenerf: A style-based 3d-aware generator for high-resolution image
  synthesis.
\newblock {\em arXiv preprint arXiv:2110.08985}, 2021.

\bibitem{Gupta2019LVISAD}
A.~Gupta, P.~Doll{\'a}r, and R.~B. Girshick.
\newblock Lvis: A dataset for large vocabulary instance segmentation.
\newblock {\em 2019 IEEE/CVF Conference on Computer Vision and Pattern
  Recognition (CVPR)}, pages 5351--5359, 2019.

\bibitem{gupta20233dgen}
A.~Gupta, W.~Xiong, Y.~Nie, I.~Jones, and B.~O{\u{g}}uz.
\newblock 3dgen: Triplane latent diffusion for textured mesh generation.
\newblock {\em arXiv preprint arXiv:2303.05371}, 2023.

\bibitem{Gupta2022VisualPC}
T.~Gupta and A.~Kembhavi.
\newblock Visual programming: Compositional visual reasoning without training.
\newblock {\em ArXiv}, abs/2211.11559, 2022.

\bibitem{hertz2022prompt}
A.~Hertz, R.~Mokady, J.~Tenenbaum, K.~Aberman, Y.~Pritch, and D.~Cohen-Or.
\newblock Prompt-to-prompt image editing with cross attention control.
\newblock {\em arXiv preprint arXiv:2208.01626}, 2022.

\bibitem{hertzmann2001image}
A.~Hertzmann, C.~E. Jacobs, N.~Oliver, B.~Curless, and D.~H. Salesin.
\newblock Image analogies.
\newblock In {\em Proceedings of the 28th annual conference on Computer
  graphics and interactive techniques}, pages 327--340, 2001.

\bibitem{ho2020denoising}
J.~Ho, A.~Jain, and P.~Abbeel.
\newblock Denoising diffusion probabilistic models.
\newblock {\em Advances in Neural Information Processing Systems},
  33:6840--6851, 2020.

\bibitem{hu2023tifa}
Y.~Hu, B.~Liu, J.~Kasai, Y.~Wang, M.~Ostendorf, R.~Krishna, and N.~A. Smith.
\newblock Tifa: Accurate and interpretable text-to-image faithfulness
  evaluation with question answering.
\newblock {\em arXiv preprint arXiv:2303.11897}, 2023.

\bibitem{johnson2017clevr}
J.~Johnson, B.~Hariharan, L.~Van Der~Maaten, L.~Fei-Fei, C.~Lawrence~Zitnick,
  and R.~Girshick.
\newblock Clevr: A diagnostic dataset for compositional language and elementary
  visual reasoning.
\newblock In {\em Proceedings of the IEEE conference on computer vision and
  pattern recognition}, pages 2901--2910, 2017.

\bibitem{kang2023scaling}
M.~Kang, J.-Y. Zhu, R.~Zhang, J.~Park, E.~Shechtman, S.~Paris, and T.~Park.
\newblock Scaling up gans for text-to-image synthesis.
\newblock {\em arXiv preprint arXiv:2303.05511}, 2023.

\bibitem{karras2021alias}
T.~Karras, M.~Aittala, S.~Laine, E.~H{\"a}rk{\"o}nen, J.~Hellsten, J.~Lehtinen,
  and T.~Aila.
\newblock Alias-free generative adversarial networks.
\newblock {\em Advances in Neural Information Processing Systems}, 34, 2021.

\bibitem{karras2019style}
T.~Karras, S.~Laine, and T.~Aila.
\newblock A style-based generator architecture for generative adversarial
  networks.
\newblock In {\em Proceedings of the IEEE/CVF Conference on Computer Vision and
  Pattern Recognition}, 2019.

\bibitem{karras2020analyzing}
T.~Karras, S.~Laine, M.~Aittala, J.~Hellsten, J.~Lehtinen, and T.~Aila.
\newblock Analyzing and improving the image quality of stylegan.
\newblock In {\em Proceedings of the IEEE/CVF Conference on Computer Vision and
  Pattern Recognition}, 2020.

\bibitem{kawar2023imagic}
B.~Kawar, S.~Zada, O.~Lang, O.~Tov, H.~Chang, T.~Dekel, I.~Mosseri, and
  M.~Irani.
\newblock Imagic: Text-based real image editing with diffusion models.
\newblock In {\em Conference on Computer Vision and Pattern Recognition 2023},
  2023.

\bibitem{king2016rearrangement}
J.~E. King, M.~Cognetti, and S.~S. Srinivasa.
\newblock Rearrangement planning using object-centric and robot-centric action
  spaces.
\newblock In {\em 2016 IEEE International Conference on Robotics and Automation
  (ICRA)}, pages 3940--3947. IEEE, 2016.

\bibitem{Kirillov2023SegmentA}
A.~Kirillov, E.~Mintun, N.~Ravi, H.~Mao, C.~Rolland, L.~Gustafson, T.~Xiao,
  S.~Whitehead, A.~C. Berg, W.-Y. Lo, P.~Doll{\'a}r, and R.~B. Girshick.
\newblock Segment anything.
\newblock {\em ArXiv}, abs/2304.02643, 2023.

\bibitem{kolve2017ai2}
E.~Kolve, R.~Mottaghi, W.~Han, E.~VanderBilt, L.~Weihs, A.~Herrasti, M.~Deitke,
  K.~Ehsani, D.~Gordon, Y.~Zhu, et~al.
\newblock Ai2-thor: An interactive 3d environment for visual ai.
\newblock {\em arXiv preprint arXiv:1712.05474}, 2017.

\bibitem{kolve2017ai2thor}
E.~Kolve, R.~Mottaghi, W.~Han, E.~VanderBilt, L.~Weihs, A.~Herrasti, M.~Deitke,
  K.~Ehsani, D.~Gordon, Y.~Zhu, A.~Kembhavi, A.~K. Gupta, and A.~Farhadi.
\newblock Ai2-thor: An interactive 3d environment for visual ai.
\newblock {\em arXiv e-prints}, pages arXiv--1712, 2017.

\bibitem{kumari2022customdiffusion}
N.~Kumari, B.~Zhang, R.~Zhang, E.~Shechtman, and J.-Y. Zhu.
\newblock Multi-concept customization of text-to-image diffusion.
\newblock In {\em CVPR}, 2023.

\bibitem{imcondsd}
LambdaLabs.
\newblock {https://lambdalabs.com/}.
\newblock
  \url{https://huggingface.co/lambdalabs/stable-diffusion-image-conditioned},
  2023.
\newblock [Online; accessed 17-May-2023].

\bibitem{laput2013pixeltone}
G.~P. Laput, M.~Dontcheva, G.~Wilensky, W.~Chang, A.~Agarwala, J.~Linder, and
  E.~Adar.
\newblock Pixeltone: A multimodal interface for image editing.
\newblock In {\em Proceedings of the SIGCHI Conference on Human Factors in
  Computing Systems}, pages 2185--2194, 2013.

\bibitem{liao2012subdivision}
Z.~Liao, H.~Hoppe, D.~Forsyth, and Y.~Yu.
\newblock A subdivision-based representation for vector image editing.
\newblock {\em IEEE transactions on visualization and computer graphics}, 2012.

\bibitem{liu2023zero}
R.~Liu, R.~Wu, B.~Van~Hoorick, P.~Tokmakov, S.~Zakharov, and C.~Vondrick.
\newblock Zero-1-to-3: Zero-shot one image to 3d object.
\newblock {\em arXiv preprint arXiv:2303.11328}, 2023.

\bibitem{liu2022structformer}
W.~Liu, C.~Paxton, T.~Hermans, and D.~Fox.
\newblock Structformer: Learning spatial structure for language-guided semantic
  rearrangement of novel objects.
\newblock In {\em 2022 International Conference on Robotics and Automation
  (ICRA)}, pages 6322--6329. IEEE, 2022.

\bibitem{Minderer2022SimpleOO}
M.~Minderer, A.~A. Gritsenko, A.~Stone, M.~Neumann, D.~Weissenborn,
  A.~Dosovitskiy, A.~Mahendran, A.~Arnab, M.~Dehghani, Z.~Shen, X.~Wang,
  X.~Zhai, T.~Kipf, and N.~Houlsby.
\newblock Simple open-vocabulary object detection with vision transformers.
\newblock {\em ArXiv}, abs/2205.06230, 2022.

\bibitem{mou2023t2i}
C.~Mou, X.~Wang, L.~Xie, J.~Zhang, Z.~Qi, Y.~Shan, and X.~Qie.
\newblock T2i-adapter: Learning adapters to dig out more controllable ability
  for text-to-image diffusion models.
\newblock {\em arXiv preprint arXiv:2302.08453}, 2023.

\bibitem{nichol2021glide}
A.~Nichol, P.~Dhariwal, A.~Ramesh, P.~Shyam, P.~Mishkin, B.~McGrew,
  I.~Sutskever, and M.~Chen.
\newblock Glide: Towards photorealistic image generation and editing with
  text-guided diffusion models.
\newblock {\em arXiv preprint arXiv:2112.10741}, 2021.

\bibitem{nichol2021improved}
A.~Q. Nichol and P.~Dhariwal.
\newblock Improved denoising diffusion probabilistic models.
\newblock In {\em International Conference on Machine Learning}, pages
  8162--8171. PMLR, 2021.

\bibitem{park2019semantic}
T.~Park, M.-Y. Liu, T.-C. Wang, and J.-Y. Zhu.
\newblock Semantic image synthesis with spatially-adaptive normalization.
\newblock In {\em Proceedings of the IEEE/CVF Conference on Computer Vision and
  Pattern Recognition}, 2019.

\bibitem{poole2022dreamfusion}
B.~Poole, A.~Jain, J.~T. Barron, and B.~Mildenhall.
\newblock Dreamfusion: Text-to-3d using 2d diffusion.
\newblock {\em arXiv preprint arXiv:2209.14988}, 2022.

\bibitem{qureshi2021nerp}
A.~H. Qureshi, A.~Mousavian, C.~Paxton, M.~C. Yip, and D.~Fox.
\newblock Nerp: Neural rearrangement planning for unknown objects.
\newblock {\em arXiv preprint arXiv:2106.01352}, 2021.

\bibitem{radford2021learning}
A.~Radford, J.~W. Kim, C.~Hallacy, A.~Ramesh, G.~Goh, S.~Agarwal, G.~Sastry,
  A.~Askell, P.~Mishkin, J.~Clark, et~al.
\newblock Learning transferable visual models from natural language
  supervision.
\newblock In {\em International conference on machine learning}, pages
  8748--8763. PMLR, 2021.

\bibitem{ramesh2022hierarchical}
A.~Ramesh, P.~Dhariwal, A.~Nichol, C.~Chu, and M.~Chen.
\newblock Hierarchical text-conditional image generation with clip latents.
\newblock {\em arXiv preprint arXiv:2204.06125}, 2022.

\bibitem{ramesh2021zero}
A.~Ramesh, M.~Pavlov, G.~Goh, S.~Gray, C.~Voss, A.~Radford, M.~Chen, and
  I.~Sutskever.
\newblock Zero-shot text-to-image generation.
\newblock In {\em International Conference on Machine Learning}, pages
  8821--8831. PMLR, 2021.

\bibitem{reinhard2001color}
E.~Reinhard, M.~Adhikhmin, B.~Gooch, and P.~Shirley.
\newblock Color transfer between images.
\newblock {\em IEEE Computer graphics and applications}, 21(5):34--41, 2001.

\bibitem{richardson2021encoding}
E.~Richardson, Y.~Alaluf, O.~Patashnik, Y.~Nitzan, Y.~Azar, S.~Shapiro, and
  D.~Cohen-Or.
\newblock Encoding in style: a stylegan encoder for image-to-image translation.
\newblock In {\em Proceedings of the IEEE/CVF Conference on Computer Vision and
  Pattern Recognition}, pages 2287--2296, 2021.

\bibitem{rombach2021highresolution}
R.~Rombach, A.~Blattmann, D.~Lorenz, P.~Esser, and B.~Ommer.
\newblock High-resolution image synthesis with latent diffusion models, 2021.

\bibitem{rombach2022high}
R.~Rombach, A.~Blattmann, D.~Lorenz, P.~Esser, and B.~Ommer.
\newblock High-resolution image synthesis with latent diffusion models.
\newblock In {\em Proceedings of the IEEE/CVF Conference on Computer Vision and
  Pattern Recognition}, pages 10684--10695, 2022.

\bibitem{saharia2022photorealistic}
C.~Saharia, W.~Chan, S.~Saxena, L.~Li, J.~Whang, E.~L. Denton, K.~Ghasemipour,
  R.~Gontijo~Lopes, B.~Karagol~Ayan, T.~Salimans, et~al.
\newblock Photorealistic text-to-image diffusion models with deep language
  understanding.
\newblock {\em Advances in Neural Information Processing Systems},
  35:36479--36494, 2022.

\bibitem{schuhmann2022laion}
C.~Schuhmann, R.~Beaumont, R.~Vencu, C.~Gordon, R.~Wightman, M.~Cherti,
  T.~Coombes, A.~Katta, C.~Mullis, M.~Wortsman, et~al.
\newblock Laion-5b: An open large-scale dataset for training next generation
  image-text models.
\newblock {\em arXiv preprint arXiv:2210.08402}, 2022.

\bibitem{schuhmann2022laion5b}
C.~Schuhmann, R.~Beaumont, R.~Vencu, C.~Gordon, R.~Wightman, M.~Cherti,
  T.~Coombes, A.~Katta, C.~Mullis, M.~Wortsman, P.~Schramowski, S.~Kundurthy,
  K.~Crowson, L.~Schmidt, R.~Kaczmarczyk, and J.~Jitsev.
\newblock Laion-5b: An open large-scale dataset for training next generation
  image-text models, 2022.

\bibitem{shen2020interfacegan}
Y.~Shen, C.~Yang, X.~Tang, and B.~Zhou.
\newblock Interfacegan: Interpreting the disentangled face representation
  learned by gans.
\newblock {\em IEEE transactions on pattern analysis and machine intelligence},
  2020.

\bibitem{shoshan2021gan}
A.~Shoshan, N.~Bhonker, I.~Kviatkovsky, and G.~Medioni.
\newblock Gan-control: Explicitly controllable gans.
\newblock In {\em Proceedings of the IEEE/CVF International Conference on
  Computer Vision}, pages 14083--14093, 2021.

\bibitem{song2020denoising}
J.~Song, C.~Meng, and S.~Ermon.
\newblock Denoising diffusion implicit models.
\newblock {\em arXiv preprint arXiv:2010.02502}, 2020.

\bibitem{szot2021habitat}
A.~Szot, A.~Clegg, E.~Undersander, E.~Wijmans, Y.~Zhao, J.~Turner, N.~Maestre,
  M.~Mukadam, D.~Chaplot, O.~Maksymets, A.~Gokaslan, V.~Vondrus, S.~Dharur,
  F.~Meier, W.~Galuba, A.~Chang, Z.~Kira, V.~Koltun, J.~Malik, M.~Savva, and
  D.~Batra.
\newblock Habitat 2.0: Training home assistants to rearrange their habitat.
\newblock In {\em Advances in Neural Information Processing Systems (NeurIPS)},
  2021.

\bibitem{tang2023make}
J.~Tang, T.~Wang, B.~Zhang, T.~Zhang, R.~Yi, L.~Ma, and D.~Chen.
\newblock Make-it-3d: High-fidelity 3d creation from a single image with
  diffusion prior.
\newblock {\em arXiv preprint arXiv:2303.14184}, 2023.

\bibitem{ulyanov2018deep}
D.~Ulyanov, A.~Vedaldi, and V.~Lempitsky.
\newblock Deep image prior.
\newblock In {\em Proceedings of the IEEE Conference on Computer Vision and
  Pattern Recognition}, 2018.

\bibitem{wang2022score}
H.~Wang, X.~Du, J.~Li, R.~A. Yeh, and G.~Shakhnarovich.
\newblock Score jacobian chaining: Lifting pretrained 2d diffusion models for
  3d generation.
\newblock {\em arXiv preprint arXiv:2212.00774}, 2022.

\bibitem{wang2020scenem}
H.~Wang, W.~Liang, and L.-F. Yu.
\newblock Scene mover: Automatic move planning for scene arrangement by deep
  reinforcement learning.
\newblock {\em ACM Transactions on Graphics}, 39(6), 2020.

\bibitem{wangcadsim}
J.~Wang, S.~Manivasagam, Y.~Chen, Z.~Yang, I.~A. B{\^a}rsan, A.~J. Yang, W.-C.
  Ma, and R.~Urtasun.
\newblock Cadsim: Robust and scalable in-the-wild 3d reconstruction for
  controllable sensor simulation.
\newblock In {\em 6th Annual Conference on Robot Learning}, 2022.

\bibitem{wang2020people}
Y.~Wang, B.~L. Curless, and S.~M. Seitz.
\newblock People as scene probes.
\newblock In {\em European Conference on Computer Vision}, pages 438--454.
  Springer, 2020.

\bibitem{wang2021repopulating}
Y.~Wang, A.~Liu, R.~Tucker, J.~Wu, B.~L. Curless, S.~M. Seitz, and N.~Snavely.
\newblock Repopulating street scenes.
\newblock In {\em Proceedings of the IEEE/CVF Conference on Computer Vision and
  Pattern Recognition}, 2021.

\bibitem{wei2023lego}
Q.~A. Wei, S.~Ding, J.~J. Park, R.~Sajnani, A.~Poulenard, S.~Sridhar, and
  L.~Guibas.
\newblock Lego-net: Learning regular rearrangements of objects in rooms.
\newblock {\em arXiv preprint arXiv:2301.09629}, 2023.

\bibitem{weihs2021visual}
L.~Weihs, M.~Deitke, A.~Kembhavi, and R.~Mottaghi.
\newblock Visual room rearrangement.
\newblock In {\em Proceedings of the IEEE/CVF conference on computer vision and
  pattern recognition}, pages 5922--5931, 2021.

\bibitem{xiazamirhe2018gibsonenv}
F.~Xia, A.~R.~Zamir, Z.-Y. He, A.~Sax, J.~Malik, and S.~Savarese.
\newblock Gibson env: real-world perception for embodied agents.
\newblock In {\em Computer Vision and Pattern Recognition (CVPR), 2018 IEEE
  Conference on}. IEEE, 2018.

\bibitem{yi2019clevrer}
K.~Yi, C.~Gan, Y.~Li, P.~Kohli, J.~Wu, A.~Torralba, and J.~B. Tenenbaum.
\newblock Clevrer: Collision events for video representation and reasoning.
\newblock {\em arXiv preprint arXiv:1910.01442}, 2019.

\bibitem{Zeng2022SocraticMC}
A.~Zeng, A.~S. Wong, S.~Welker, K.~Choromanski, F.~Tombari, A.~Purohit, M.~S.
  Ryoo, V.~Sindhwani, J.~Lee, V.~Vanhoucke, and P.~R. Florence.
\newblock Socratic models: Composing zero-shot multimodal reasoning with
  language.
\newblock {\em ArXiv}, abs/2204.00598, 2022.

\bibitem{zhang2023adding}
L.~Zhang and M.~Agrawala.
\newblock Adding conditional control to text-to-image diffusion models, 2023.

\bibitem{zhu2020indomain}
J.~Zhu, Y.~Shen, D.~Zhao, and B.~Zhou.
\newblock In-domain gan inversion for real image editing.
\newblock In {\em Proceedings of European Conference on Computer Vision
  (ECCV)}, 2020.

\bibitem{zhu2017unpaired}
J.-Y. Zhu, T.~Park, P.~Isola, and A.~A. Efros.
\newblock Unpaired image-to-image translation using cycle-consistent
  adversarial networks.
\newblock In {\em Proceedings of the IEEE international conference on computer
  vision}, 2017.

\end{thebibliography}
\bibliographystyle{abbrv}

\label{sec:supplementary}

\section{Appendix}

\subsection{Training and Inference Details}

We closely follow the training procedure established by \cite{liu2023zero}, with a few modifications. Our approach uses an effective batch size of $1024$, which is smaller than the batch size of $1536$ used by \zero. This adjustment was necessary because of the additional memory requirements caused by the reintroduction of the CLIP text encoder. This batch size is achieved by using a local batch size of $64$ across 40GB NVIDIA RTX A6000 GPUs, along with two gradient accumulation steps. Similar to Zero123, we train on images with a resolution of $256\times 256$, resulting in a latent spatial dimension of $32\times 32$. Following their protocol, we utilize the AdamW optimizer, with a learning rate of 1e-4 for all parameters of the model except for those of the concatenation MLP, which uses a learning rate of $1e-3$. Our training process runs for a total of 20,000 steps. We then select the best checkpoint based on our metrics computed from an unseen object validation set. As was the case in StableDiffusion, we freeze the CLIP text encoder during training. For inference, we generate images with the DDIM~\cite{song2020denoising} sampler using 200 steps. We do not use classifier-free guidance, i.e. the cfg term is set to $1.0$.

\subsection{Robustness to Severity of Transformation}

\begin{figure}[ht]
  \centering
  \begin{minipage}[b]{0.45\textwidth}
    \includegraphics[width=\textwidth]{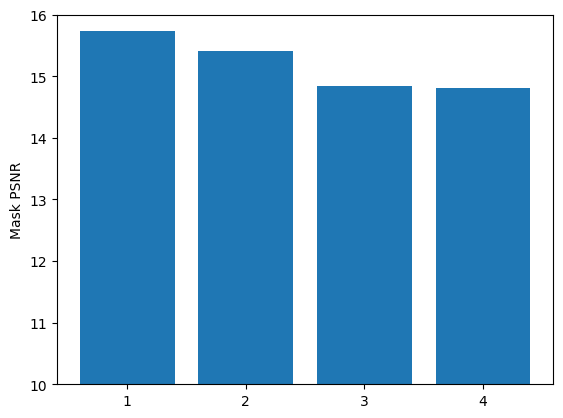}
    \caption{Average Mask PSNR of the single-task rotation model as the number of objects in the scene varies.}
    \label{fig:clutter}
  \end{minipage}
  \hfill
  \begin{minipage}[b]{0.45\textwidth}
    \includegraphics[width=\textwidth]{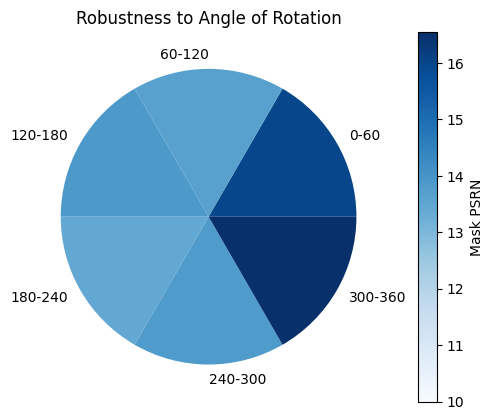}
    \caption{Average Mask PSNR of the single-task rotation model for rotation angles falling within a slice. }
    \label{fig:angle_metric}
  \end{minipage}
\end{figure}

We analyze the robustness of our method by measuring the performance of the single task rotation model as the complexity of the scene and severity of transformation changes. In Figure~\ref{fig:clutter}, we show the average of our Mask PSNR metric as the number of objects in the scene varies from $1$ to $4$, where a slight drop in performance occurs as the number of objects increases. In Figure~\ref{fig:angle_metric}, we show average Mask PSNR for rotations in a given angle range on a pie chart, where it can be seen that the model does better with smaller angle deviations.

\subsection{Additional Results}

\begin{figure}[ht]
  \centering
  \includegraphics[width=\textwidth]{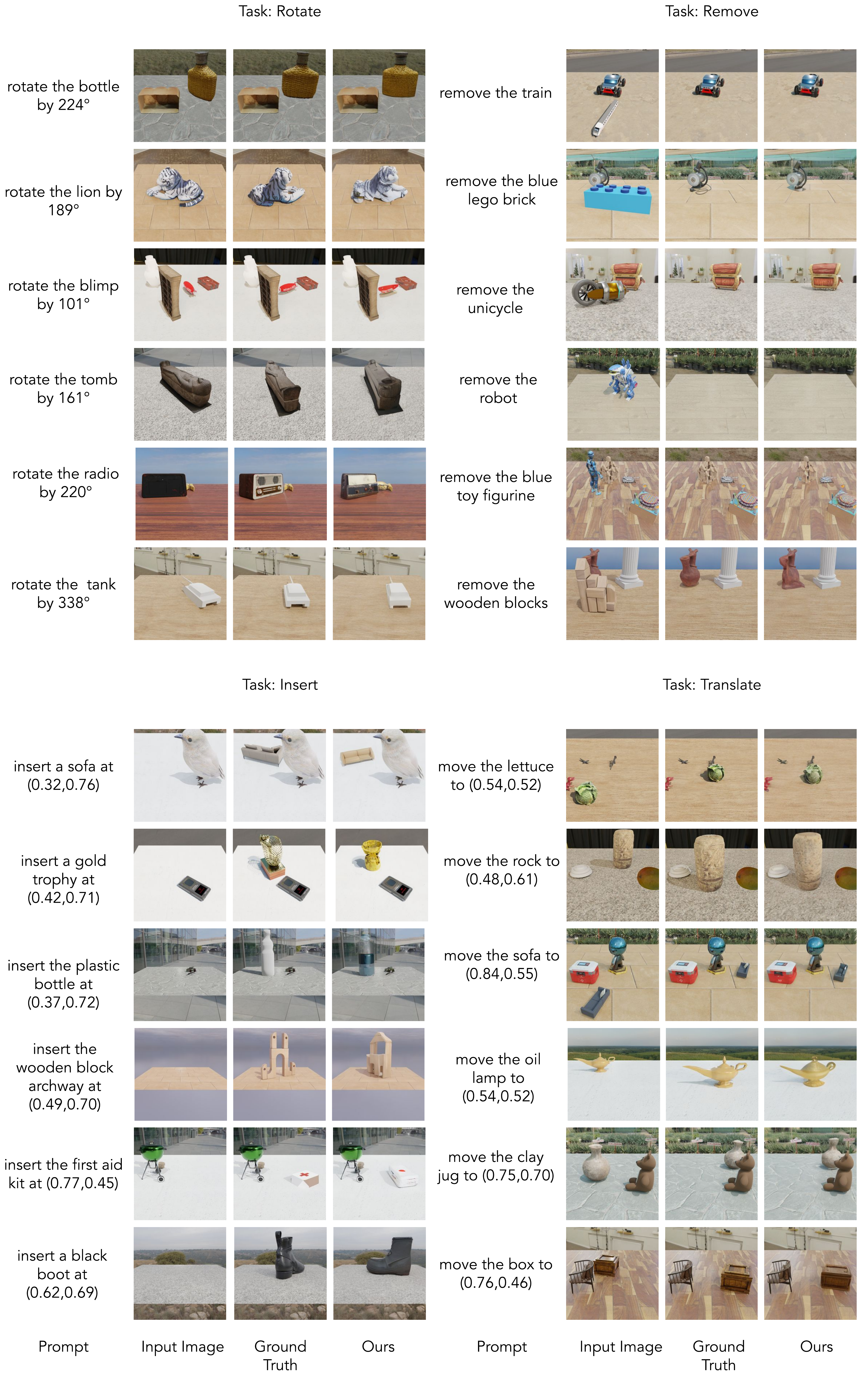}
  \caption{Generated examples by the \model\ multitask model.}
  \label{fig:sup_multi}
  \vspace{-1em}
\end{figure}

In Figure~\ref{fig:sup_multi}, we show qualitative results from our multitask model on each of the four editing tasks.

\subsection{Dataset Analysis}

\begin{figure}[ht]
  \centering
  \begin{minipage}[b]{0.45\textwidth}
    \includegraphics[width=\textwidth]{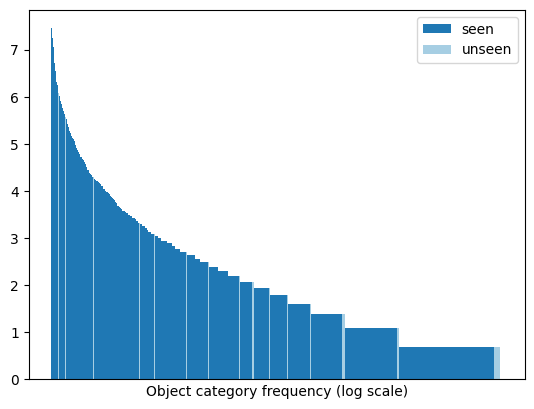}
    \caption{Object categories from seen and unseen splits sorted by frequency, in log scale.}
    \label{fig:cat_hist}
  \end{minipage}
  \hfill
  \begin{minipage}[b]{0.45\textwidth}
    \includegraphics[width=\textwidth]{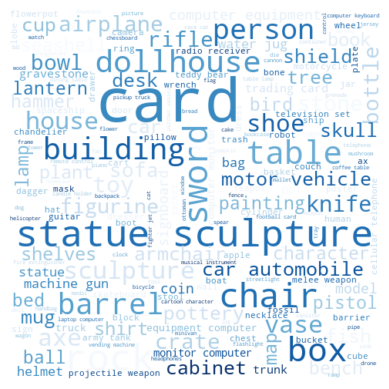}
    \caption{A wordcloud visualizing the frequency of various object category names.}
    \label{fig:word_cloud}
  \end{minipage}
\end{figure}

In this seciton, we provide some details about the composition and statistical makeup of our dataset. In Table~\ref{tab:dataset-stat}, we show a statistical overview of the dataset, including total number of objects and categories, as well as the mean, median, and standard deviation of objects per category. We also visualize the distribution of objects across categories, as can be show in the histogram in Figure~\ref{fig:cat_hist}. Finally, we visualize the frequency of category names in the wordcloud in Figure~\ref{fig:word_cloud}.

\begin{table}[]
\centering
\caption{Summary of key statistics of the \dataset\ dataset.}
\label{tab:dataset-stat}
\begin{tabular}{ll}
\hline
Total objects              & 62950  \\
Total categories           & 1613   \\
Object per category median & 6.0    \\
Object per category mean   & 39.03  \\
Object per category std    & 138.00 \\ \hline
\end{tabular}
\end{table}

\end{document}